\documentclass[letterpaper]{article} 
\usepackage{aaai2026}  
\usepackage{times}  
\usepackage{helvet}  
\usepackage{courier}  
\usepackage[hyphens]{url}  
\usepackage{graphicx} 
\usepackage{natbib}  
\usepackage{caption} 
\usepackage{algorithm}
\usepackage{algorithmic}

\usepackage{newfloat}
\usepackage{listings}
\DeclareCaptionStyle{ruled}{labelfont=normalfont,labelsep=colon,strut=off} 
\floatstyle{ruled}
\newfloat{listing}{tb}{lst}{}
\floatname{listing}{Listing}
\title{A Visualized Framework for Event Cooperation with Generative Agents}

\author{
    Yuyang Tian \textsuperscript{\rm 1,2}\footnotemark[1]\quad 
    Shunqiang Mao \textsuperscript{\rm 1,3}\footnotemark[1]\quad
    Wenchang Gao \textsuperscript{\rm 1}\quad
    Lanlan Qiu \textsuperscript{\rm 1}\quad
    Tianxing He \textsuperscript{\rm 4, 1}\footnotemark[2]
}
\affiliations{
    \textsuperscript{\rm 1}Shanghai Qi Zhi Institute, 
    \textsuperscript{\rm 2}University of Science and Technology of China, 
    \textsuperscript{\rm 3}Sun Yat-sen University, 
    \textsuperscript{\rm 4}Tsinghua University, \\
    yytian2004@gmail.com, 
    1400126712@qq.com,
    gaowenchang@sqz.ac.cn,
    qiulanlan@sqz.ac.cn,
    hetianxing@mail.tsinghua.edu.cn
}

\usepackage{bibentry}

\begin{document}

\maketitle

\begin{abstract}
Large Language Models (LLMs) have revolutionized the simulation of agent societies, enabling autonomous planning, memory formation, and social interactions. However, existing frameworks often overlook systematic evaluations for event organization and lack visualized integration with physically grounded environments, limiting agents' ability to navigate spaces and interact with items realistically. We develop MiniAgentPro, a visualization platform featuring an intuitive map editor for customizing environments and a simulation player with smooth animations. Based on this tool, we introduce a comprehensive test set comprising eight diverse event scenarios with basic and hard variants to assess agents' ability. Evaluations using GPT-4o demonstrate strong performance in basic settings but highlight coordination challenges in hard variants.

\end{abstract}

\section{Introduction}

\def\thefootnote{*}\footnotetext{Equal Contribution} \def\thefootnote{\arabic{footnote}}

\def\thefootnote{\dag}\footnotetext{Corresponding Author} \def\thefootnote{\arabic{footnote}}

The landscape of artificial intelligence has been profoundly reshaped by recent breakthroughs in Large Language Models (LLMs) \citep{hong2024metagpt, Singhal2023}. These powerful models have demonstrated an unprecedented ability to achieve human-level performance across a multitude of tasks, ranging from complex reasoning to creative content generation \citep{chen-etal-2025-towards-medical, zhou2023leasttomost}. This capability has, in turn, infused new vitality into the concept of building agent societies. Generative Agent (GA) \citep{park} demonstrates how LLM-driven agents could autonomously plan, form memories, and engage in believable interactions within a sandbox environment. Subsequent research has focused on enriching the generative agent paradigm by integrating more sophisticated and human-centric elements \citep{li2024evolvingagentsinteractivesimulation,humanoidagents}. Additionally, efforts have extended to downstream applications, including policy simulation \citep{hou2025societygenerativeagentssimulate, park2024generativeagentsimulations1000}, sociological theory modeling \citep{yang2025agentexchangeshapingfuture}, and the design of non-player characters (NPCs) in games \citep{Exploring_Presence, Conversational_Interactions}.

However, there's an absence of systematic test sets to evaluate how well these agents perform in organizing various events, such as a birthday party, in addition to their routine daily plans. Furthermore, current simulations often lack physical grounding, which inherently limits the evaluation of agents' ability to exhibit exquisite interactions with the simulation environment. 


Our work makes the following contributions:

\begin{itemize}
    

\item We develop a visualization platform for simulation play, paired with a map editor, to streamline human evaluation and enable easy customization of environments for future research.

\item We enhance the prior GA framework with on-the-fly planning, basic physics, and item interactions for more realistic simulations. We also introduce a test set and evaluation protocol across 8 diverse task settings to assess agents' event coordination abilities.
    
\end{itemize}

\section{Visualization Tool}

\begin{figure*}[t]
    \centering
    \begin{minipage}[t]{0.495\linewidth}
        \centering
        \includegraphics[width=\linewidth]{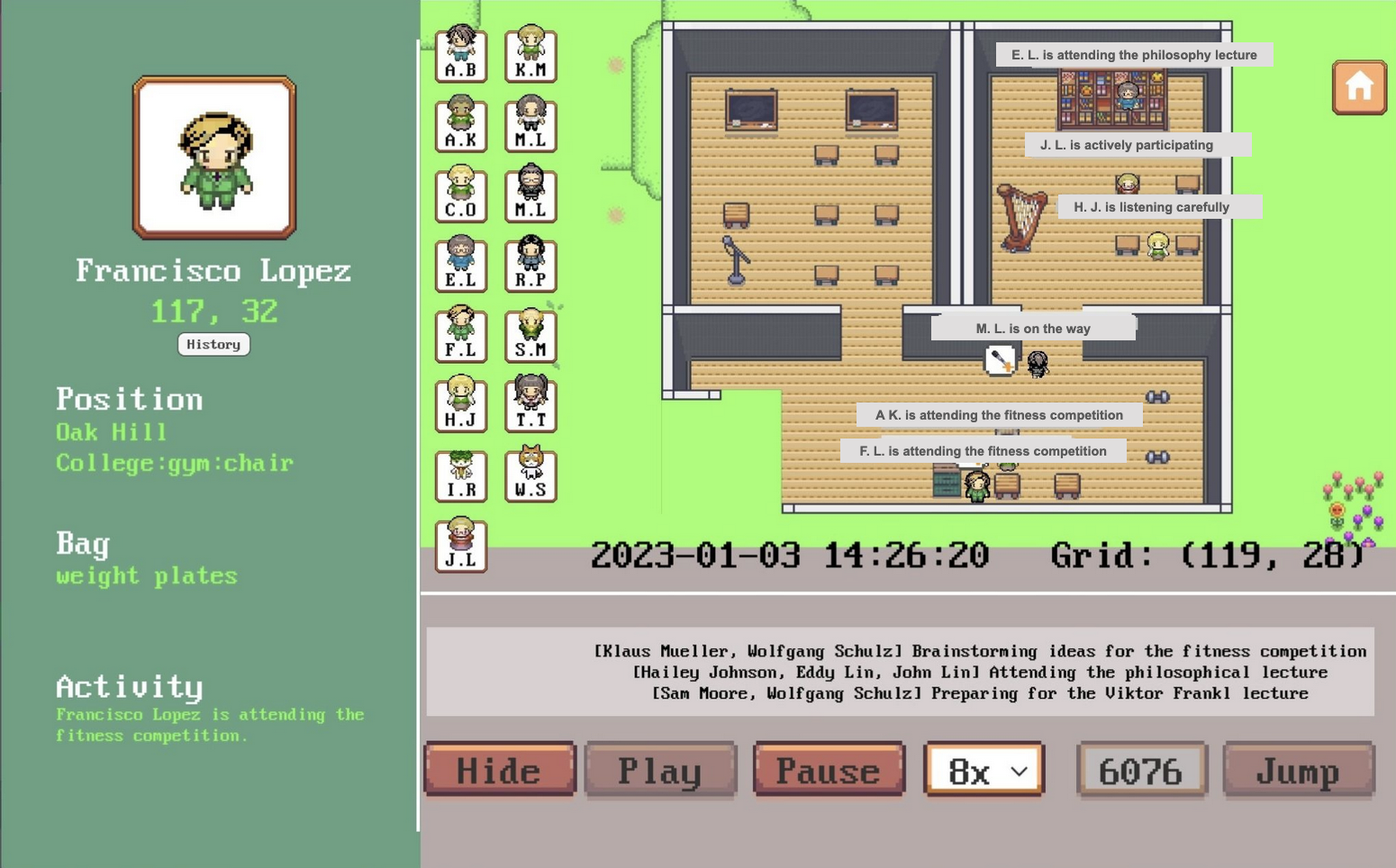}
        \label{fig:player}
    \end{minipage}
    \hfill
    \begin{minipage}[t]{0.495\linewidth}
        \centering
        \includegraphics[width=\linewidth]{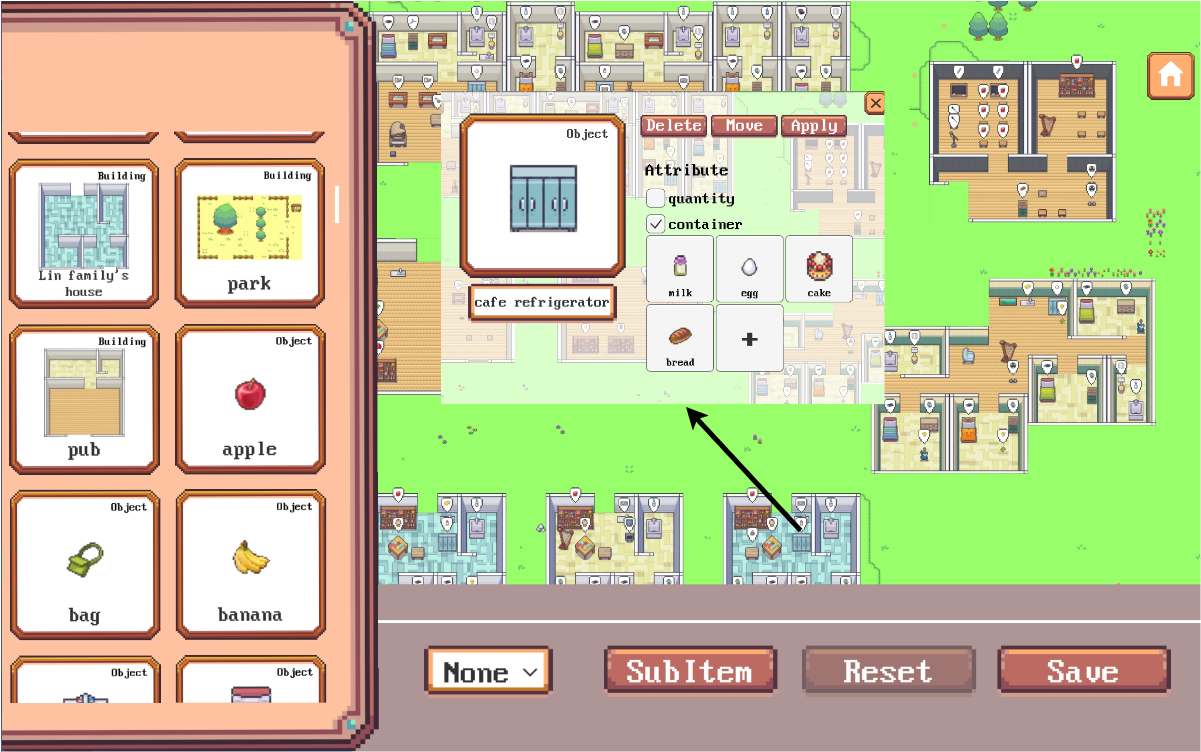}
        \label{fig:editor}
    \end{minipage}
    
    \caption{User interfaces of MiniAgentPro. (left) Simulation Player for observing agent interactions and activities. (right) Map Editor for creating and editing the environment.}
    \label{fig:player_editor}
\end{figure*}

In this section, we introduce MiniAgentPro\footnote{Our Unity executable tool is open-sourced at \\ \url{https://github.com/Just-A-Pie/MiniAgentStudio}.}, a visualization tool to improve the whole pipeline of agent simulation. Our tool contains two parts, the map editor and the simulation player as shown in Figure \ref{fig:player_editor}.


\paragraph{Map Editor}

The map editor allows users to customize the simulation environment. It offers an intuitive interface supporting a range of basic functions, including placing, moving, and deleting buildings and items. We facilitate the management of complex relationships of items through our container and property system, where items can be nested within others, and users can assign specific properties to elements, reflecting real-world scenarios. 

The editor supports environments with up to 15 agents and 100 unique types of items, providing flexibility for complex and dynamic scenarios. The editor saves the environment configuration in CSV and JSON formats, ensuring compatibility with backend simulation. 

\paragraph{Simulation Player}

We incorporate a simulation player, which allows for detailed inspection of each agent’s current state, activity history, and interactions with other agents and the map within the environment. The platform is designed to be user-friendly, with intuitive controls for playback speed, jumping to specific steps, and inspecting agent and item details. It supports smooth animation of agent movements. By leveraging Unity, we provide a visually appealing and interactive way to observe and analyze the results, making it a useful tool for researchers, educators, and developers in the field of LLM-based social simulations.

\section{Agent Framework}


Our agent framework builds on GA \citep{park}, and incorporates an on-the-fly adaptive planning mechanism, and physically grounded constraints\footnote{Our agent backend code is open-sourced at \\\url{https://github.com/tian-yuyang/EventGA}.}.

\paragraph{Activity Planning}

Following GA, we decompose an agent's daily activities into two hierarchical levels: high-level plans and low-level actions. In contrast to standard GA pipelines, where high-level plans are often predetermined at the beginning of the simulation, our system dynamically generates high-level plans and low-level actions on-the-fly as the simulation proceeds. This is achieved by leveraging each agent's unique personality descriptions, history of their past high-level plans, and the retrieved memories, so that their behaviors evolve organically in response to both internal states and external event requirements. 


Additionally, we introduce physical constraints to enhance realism. Agents must navigate the environment with a constrained movement speed, perceiving and reacting to events along the way, which introduces temporal costs and increases complexity. Furthermore, our item interaction mechanism requires agents to collect specific items associated with each low-level action before proceeding to the designated location.

\paragraph{Dialogue}

Our dialogue system, inspired by prior work on \citep{humanoidagents}, enables rich, context-aware communication that mirrors real-world interactions. Agents integrate substantial contextual information into dialogue prompts, including personality traits, core characteristics, current daily activities, life progress statements, the activities of other agents and event-related context. This information determines how a dialogue is initiated and shapes its topic, ensuring conversations are task-relevant. After that, the two agents start to generate dialogue in turn until one decides to stop. 


\section{Evaluation}

We develop a test set comprising eight diverse event settings: a fitness competition, a friends' dinner, a Lin's family party, a music jam session, a mixology workshop, an open mic comedy night, a philosophy lecture, and a writing workshop. These scenarios involve between 3 and 6 agents each, and are designed to evaluate the agents' abilities in event organization, coordination, and interaction within physically grounded environments. Each setting incorporates specific tasks, such as gathering required items, navigating to designated locations, and engaging in meaningful dialogues. We also design basic and hard versions: The basic version of each setting directly gives all participants the event information, while the hard version only informs the host about the event and lets the host invite all others.

\begin{table}[hbtp]
\centering
\resizebox{.95\columnwidth}{!}{
\begin{tabular}{c|l|l|l|l|l|l}
    \hline
    Settings & RF & LA & BIR & DRC & IQ & Avg. \\
    \hline
    basic & 7.04 & 8.69 & 6.71 & 7.87 & 6.91 & 7.44 \\
    hard & 3.21 & 4.50 & 5.14 & 8.07 & 4.50 & 5.08 \\
    \hline
\end{tabular}
}
\caption{The evaluation results of the average score from all event settings.}
\label{table1}
\end{table}

We test the criteria by self-designed metrics: Role Fulfillment (RF), Location Adherence (LA), Bag Item Relevance (BIR), Daily Requirement Consistency (DRC), and Interaction Quality (IQ). The score is given by GPT-4o mini in a range of 1 to 10, and the LLM used for simulation is GPT-4o. The results are shown in Table \ref{table1}. The evaluation reveals strong performance in basic settings, with an overall average of 7.44. In hard variants, performance drops to 5.08, with Role Fulfillment plummeting to 3.21 due to coordination failures in invitation and participation. These findings show our framework's potential for straightforward tasks but emphasize the need for advanced mechanisms to foster emergent collaboration in complex, invitation-based scenarios.

%

\bibliography{aaai2026}



\end{document}